\documentclass[wcp]{jmlr}

\usepackage{longtable}

\usepackage{booktabs}
\usepackage{multirow}
\usepackage{siunitx} 
\usepackage{graphicx} 
\usepackage{float}  
\usepackage{lineno}
\usepackage{caption} 
\usepackage{booktabs}

\makeatletter
\let\Ginclude@graphics\@org@Ginclude@graphics 
\makeatother

\jmlrvolume{222}
\jmlryear{2023}
\jmlrworkshop{ACML 2023}

\title[Learning to Terminate in Object Navigation]{Learning to Terminate in Object Navigation}

\author{\Name{Yuhang Song} \thanks{The author is affiliated with the Department of Computer Science, National Tsing Hua University as well in a duel Ph.D. program.} \Email{sgyson10@liverpool.ac.uk}\\
\addr Department of Computer Science \\
University of Liverpool
\AND
\Name{Anh Nguyen} \Email{anh.nguyen@liverpool.ac.uk}\\
\addr Department of Computer Science \\
University of Liverpool
\AND
\Name{Chun-Yi Lee} \Email{cylee@cs.nthu.edu.tw}\\
\addr Department of Computer Science \\
National Tsing Hua University
}

\editors{Berrin Yan{\i}ko\u{g}lu and Wray Buntine}

\begin{document}

\maketitle

\begin{abstract}
This paper tackles the critical challenge of object navigation in autonomous navigation systems, particularly focusing on the problem of target approach and episode termination in environments with long optimal episode length in Deep Reinforcement Learning (DRL) based methods. While effective in environment exploration and object localization, conventional DRL methods often struggle with optimal path planning and termination recognition due to a lack of depth information. To overcome these limitations, we propose a novel approach, namely the Depth-Inference Termination Agent (DITA), which incorporates a supervised model called the Judge Model to implicitly infer object-wise depth and decide termination jointly with reinforcement learning. We train our judge model along with reinforcement learning in parallel and supervise the former efficiently by reward signal. Our evaluation shows the method is demonstrating superior performance, we achieve a $9.3\%$ gain on success rate than our baseline method across all room types and gain $51.2\%$ improvements on long episodes environment while maintaining slightly better Success Weighted by Path Length (SPL). Code and resources, visualization are available at: \url{https://github.com/HuskyKingdom/DITA_acml2023}
\end{abstract}
\begin{keywords}
Visual navigation,  Supervised learning, Deep Reinforcement learning
\end{keywords}

\section{Introduction}

Object navigation represents a critical challenge within the realm of autonomous navigation \citep{bagnell2010learning}, it necessitates the ability of robotic agents to navigate proficiently within environments that have not been previously encountered. The primary goal is to reach a specified target object, and the successful completion of this task is contingent upon the agent's ability to self-declare the successful attainment of the target object, thereby concluding the episode. Such tasks may seem straightforward from a human perspective given our inherent knowledge and comprehension of the essential conditions required for successful navigation \citep{s22062387}. Humans, for example, possess an intuitive sense of where to begin exploring, as certain objects have a higher likelihood of being found in specific areas. Moreover, upon visually spotting the desired object, we instinctively plan an optimal route toward the target. Drawing inspiration from human problem-solving strategies, we could break down the task into two phases: (i) Explore the environment and locate the target object. (ii) Navigate to the target object until it is reached, then declare episode termination. 

The underlying principle of Deep Reinforcement Learning methods (DRL) of maximizing the cumulative reward, inherently aligned with the goal of effective exploration and object localization, led to their extensive use within the field. \cite{mirowski2016learning,zhu2017target} trained agents to perform navigation behaviors by encoding visual observation of the agent with its relevant states as embedding and passing that to A3C \citep{mnih2016asynchronous} Reinforcement Learning model with the recurrent neural network. \cite{wortsman2019learning} adopts a meta-learning approach with reinforcement learning, where it learns a self-supervised interaction loss during the inference process, to help prevent collisions. Moreover, By considering semantic context, just like how pre-knowledge of human beings take part, \cite{yang2018visual,pal2021learning,druon2020visual,du2020learning} propose to incorporate scene prior of the object relations with Graph Neural Network (GCN) embedded to the network for the agent to better explores the environment. 
Despite the promising outcomes demonstrated by Deep Reinforcement Learning (DRL) based methods in exploration and object localization, their application in environments characterized by extended optimal episode lengths presents distinct challenges. They often struggle to address optimal path planning to the object and termination recolonization \citep{kartal2019terminal}. In these scenarios, our observations indicate that after the agent has seen the target object, it often still fails to keep approaching the target. These limitations become even more pronounced in object navigation, where the agents are expected to declare the termination of the episode on its own in unseen environments with the absence of depth information. Given that objects of varying types often exhibit different sizes, it becomes challenging for DRL agents to discern the dependencies between their actions and the task at hand without explicit depth information pertaining to the object, resulting in the navigation agent falling into local maximums \citep{jaakkola1994reinforcement}, in which it avoids step penalty by terminating the episode in the early stage in environments \citep{electronics11213628}. 


\begin{figure}[t]
	\centering
	\includegraphics[scale=0.42]{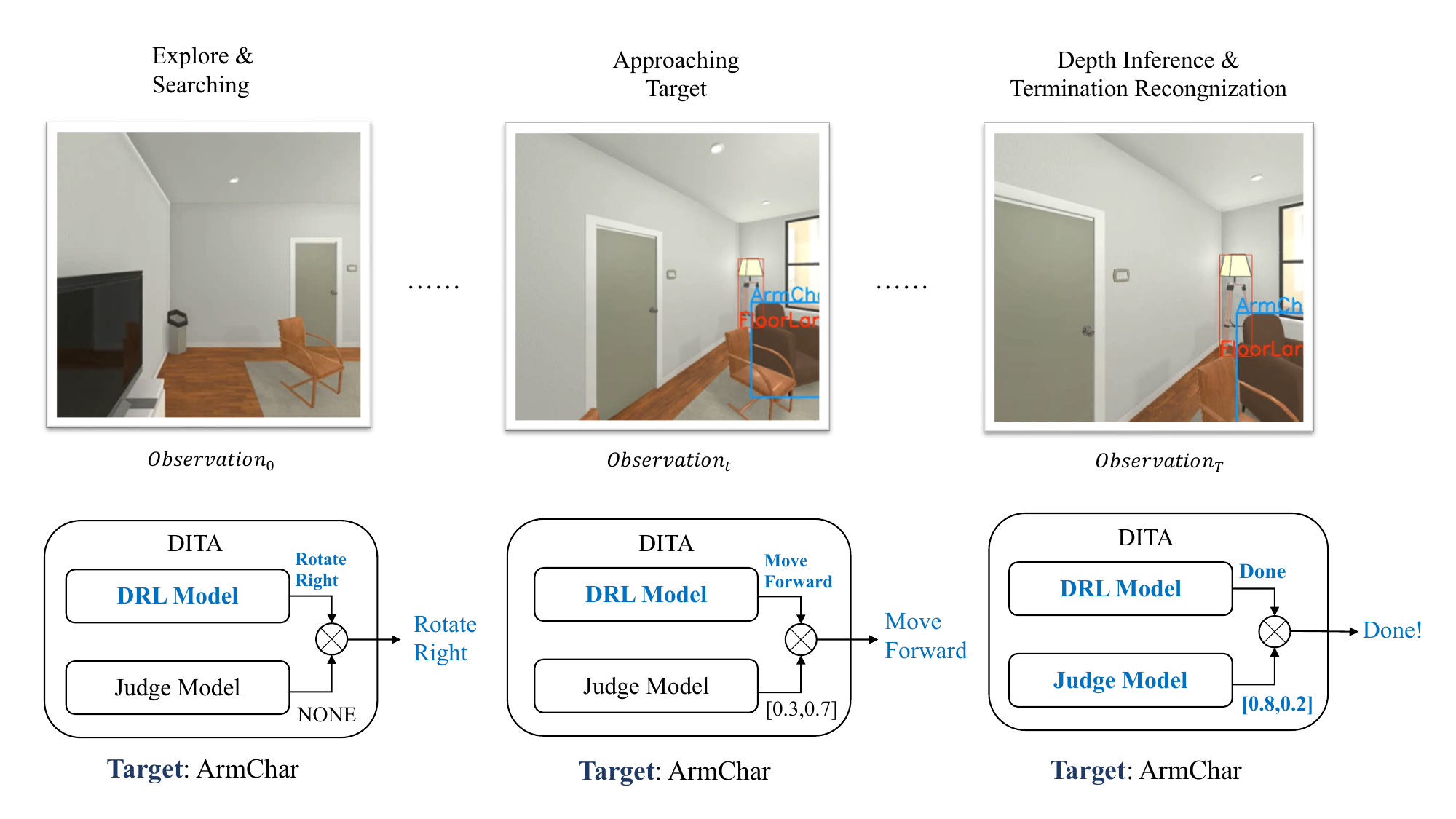}
	\caption{Depth-Inference Termination Agent (DITA) Model Overview. Upon sense observation from time step $t$, the DRL model embeds the observation into $StateEmb_{t}$, this embedding is then sent to the judge model to classify whether to sample termination action, based on both output from DRL and the judge model, our DITA model outputs the final action $a_{t}$.}
	\label{DITA_overview}
 \vspace{-10pt}
\end{figure}

Building on these insights, we introduce an innovative approach to object navigation that harnesses the power of Deep Reinforcement Learning (DRL) rewards to guide a model in inferring depth implicitly. Our method introduces a model called the \textit{Judge Model}, a supervised classification model trained in conjunction  with the DRL agent and guided by the DRL reward signal. The Judge Model's role is to assess the appropriate termination time for the DRL agent by implicitly estimating object depth based on the results of object detection. We integrate our judge model as part of the agent, enabling the DRL agent to explore the unseen environment while searching for the target. Once the target appears in the observation frame, the judge model provides a termination confidence level. The agent then decides whether to terminate the episode based on the outputs from both models as shown in Figure \ref{DITA_overview}. We evaluate our proposed DITA model in AI2-THOR framework \citep{kolve2017ai2}, a platform that furnishes highly customizable environments, and permits the agent to enact navigation actions within these environments, subsequently observing the changes induced by those actions. 

Our contributions are summarized as follows: (1) We build a supervised model called judge model to recognize termination by implicitly inference object depth. (2) The integration of the judge model with a backbone DRL, training them simultaneously. (3) Our experiment result demonstrates the generalizability of implicit depth inference to unseen environments, DITA outperforms previous pure reinforcement learning-based methods. 

The remaining of the paper is organized as the following, section \ref{sec:related work} introduces related works in the field, then we demonstrate our main approach and discuss the definition of object navigation task in \ref{sec:method}. In section \ref{sec:result} we will go through the dataset we used, with experiment designs and results, then end by section \ref{sec:conclusion} where we will summarize our work and discuss possible future works.

\begin{figure}
	\centering
	\includegraphics[scale=0.8]{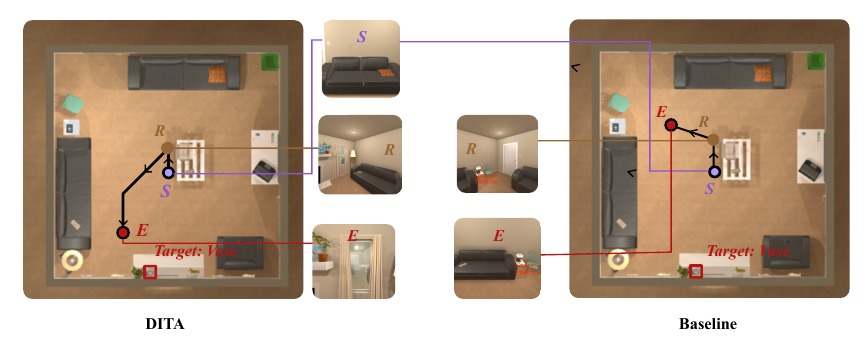}
	\caption{Trajectories of DITA and  MJOLNIR-o baseline in FloorPlan 225. Point $S$ is where the agent is initialized, $E$ is where the agent samples termination action, $R$ is where the agent rotates around to find the target. The baseline model rotates and ends the episode before it finds the target, whereas our DITA agent does not end the episode until it is confident enough.}
	\label{DITA_Traj}
    \vspace{-6pt}
\end{figure}

\section{Related Work}
\label{sec:related work}


\textbf{Map-based Navigation.} Visual Navigation refers to the tasks that with visual input for an agent to navigate. Traditional methodologies primarily focused on solving navigation problems by building explicit models of the environment in the agent's memory through interaction, enabling inference from the obtained knowledge \citep{oriolo1995line, milani2023navigates,chaplot2020learning, chaplot2020object, ramakrishnan2022poni}. This knowledge usually consists of environmental maps and additional prior knowledge. With the advent of Simultaneous Localization and Mapping (SLAM) \citep{fuentes2015visual}, a modular and hierarchical approach was proposed to construct explicit environment maps for both exploration and inference \citep{chaplot2020learning}. Subsequent studies include \cite{chaplot2020object} integrated semantic priors into the environment model, resulting in maps with semantic priors. Inferences were made on learned semantic knowledge, and a pre-trained potential function network was used to predict target potential areas from the generated top-down semantic maps \citep{ramakrishnan2022poni}. Our work deviates from these conventional approaches as our navigation model is not based on any maps, our model learns the exploration policy and target recognition simultaneously. Recently, a proposal to maintain a topological map-like Hierarchical Object-to-Zoo (HOZ) graph during navigation was made \citep{zhang2021hierarchical}, allowing agents to perform optimal path planning. However, the HOZ graph requires significant manual design and configuration, limiting its flexibility and adaptability in varied or unpredictable environments. Our approach differs by learning more generalizable implicit depth information.

\textbf{Map-less Navigation.} Due to the computational complexity and memory consumption of map-based methods, especially when constructing maps in complex environments, more attention has been directed towards map-less deep reinforcement learning models \citep{khandelwal2022simple,zhu2021soon,dinh2023habicrowd,ye2021auxiliary, 9812027}. These models usually encode the current states of the agent into an embedding and feed it into deep reinforcement learning models. These can be broadly classified into those that use more informative encoders or those based on Recurrent Neural Networks (RNN) \citep{mirowski2016learning,zhu2017target,savva2017minos,yang2018visual,pal2021learning,ramrakhya2022habitat,wijmans2023emergence}. Our work belongs to this latter category, but unlike the others, we consider estimating depth on an object-wise basis. Additionally, an alternative approach in the literature combines imitation learning with reinforcement learning frameworks \citep{du2020learning,du2021vtnet}. While the fusion of imitation and reinforcement learning presents an interesting approach, our work aims to maximize the efficiency and effectiveness of a combination of reinforcement learning and self-supervised signals. Our approach is applicable to both exploration and exploitation, even in the absence of suitable expert demonstrations.

\textbf{Problem of Local Maxima.} The issue of local maxima is a significant challenge in Reinforcement Learning. This problem, which arises from sparse rewards, hinders agents from achieving the optimal solution in complex environments with extensive action spaces. Current solutions to these problems include either leveraging existing data of the agent itself, for example, encouraging the agent to explore more on new states \citep{ostrovski2017count,pathak2017curiosity,stadie2015incentivizing, haarnoja2018soft}, or learning from states with no reward \citep{andrychowicz2017hindsight}. Alternatively by making use of external guidance, either through Reward Shaping \citep{hu2020learning,devlin2012dynamic}, Imitation Learning \citep{ho2016generative,ramrakhya2022habitat} or Curriculum Learning \citep{soviany2022curriculum}. However, these methods essentially presuppose the agent's incapacity to terminate the episode independently, which aids the exploration of diverse state possibilities in complex environments. In our context, the diverse representations of different room types and the agent's capability to enact termination action make these traditional methods less applicable or insufficiently effective. Additionally, existing exploration encouragement methods such as curiosity-driven exploration \citep{pathak2017curiosity}  might need to be adapted to ensure the agent explores not only the states but also the potential termination points effectively. Instead, we directly train a judge model alongside Reinforcement Learning to only allow the agent to actively terminate when it is confident enough.

\textbf{Depth Inference.} Depth Inference refers to the prediction of depth maps using RGB images. This area is well-established within the field of Computer Vision, as demonstrated by a plethora of studies  \citep{laina2016deeper,zhou2017unsupervised,zheng2018t2net,ranjan2019competitive}. Nonetheless, directly translating these depth estimation methodologies into our context introduces several complications. These models were initially designed either to estimate precise depth maps over the whole frame or require labeled training data in certain scenarios, direct application of these depth estimation methods into our scenarios can lead to high computational overhead or inefficiency. Conversely, our proposed method capitalizes on the results of object detection. By directly learning from the reward signal of the environment, our model implicitly infers depth information solely on specific objects of interest to determine whether to terminate the episode, making it more suitable for the task.

\section{Learning to Terminate in Object Navigation}
\label{sec:method}

\subsection{Definition of Object Navigation}
\label{sec:task}

Consider an environment set that has object types $C = \{c_1,c_2,...,c_n\}$, the aim of object navigation is to navigate to a specified object type $c_{target} \in C$,  e.g.,  an "ArmChair" or "Pillow". The agent is initially placed randomly in state $t_0$. At each time step  $t$, it takes observation $o_t$ and acts in the environment. $o_t \in O$ is a visual input of RGB image captured by the agent's camera, whereas the agent has the action space of six discrete actions $a_t \in A = \{MoveAhead, RotateLeft, RotateRight, LookUp, LookDown, Done\}$. The action $MoveAhead$ propels the agent forward $0.25\unit{\meter}$, rotational actions turn the agent $45^\circ$ to the left or right, and look actions adjust the camera by $30^\circ$ upwards or downwards. The action $Done$ enables the agent to declare success and terminate the episode. Episode termination can occur due to various conditions, including the agent's active decision to terminate or when the episode reaches its maximum predefined length. An episode is deemed successful if the agent actively terminates with the target object within the observation frame and the distance between the agent and the target object is less than $1.5\unit{\meter}$.

\begin{figure}
	\centering
	\includegraphics[scale=0.42]{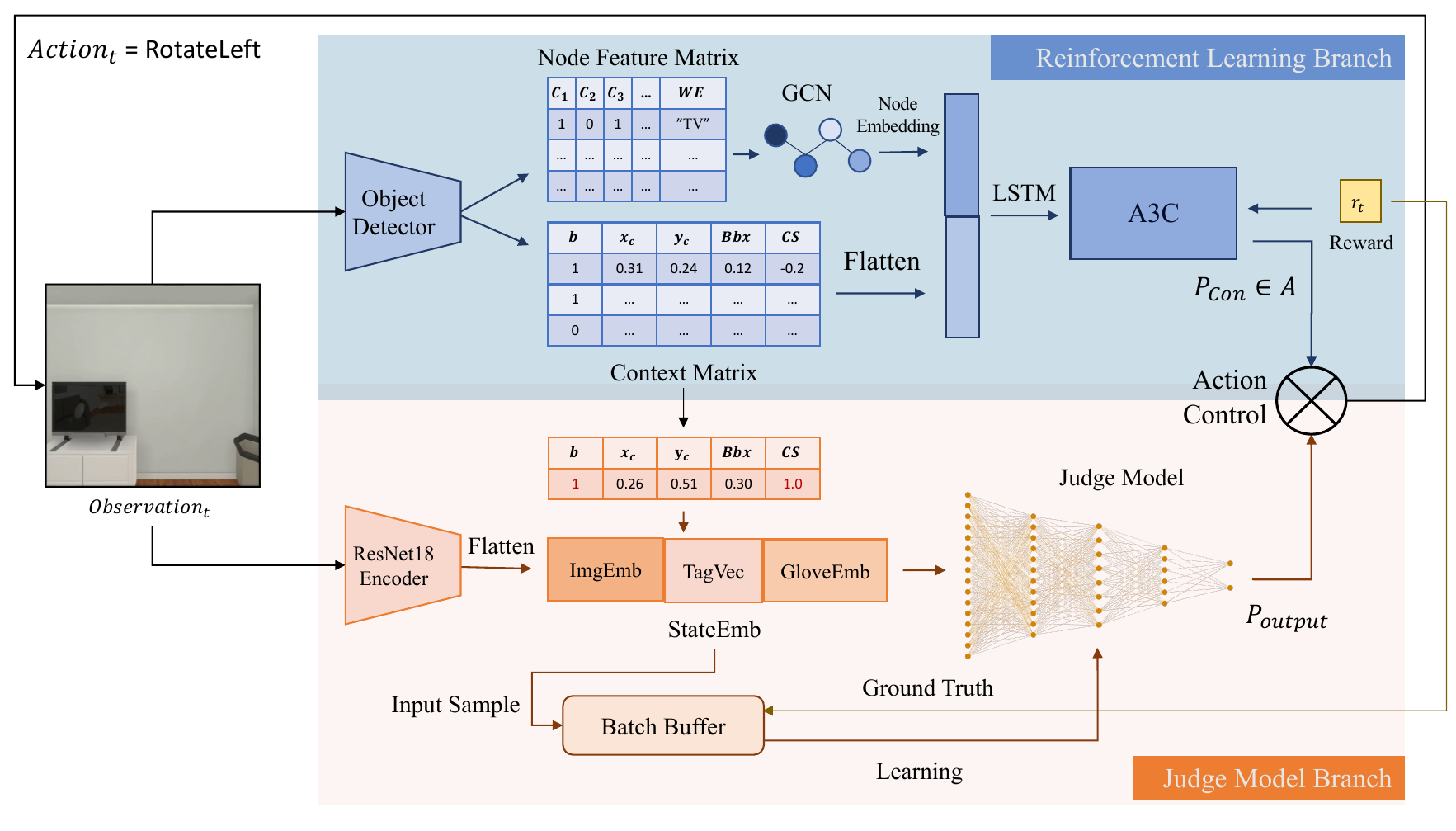}
	\caption{DITA Architecture. $Observation_t$ is passed into both the reinforcement learning branch and judge model branch, where the reinforcement learning branch outputs \textit{control action distributions} $P_{con}$, and judge model outputs termination action distribution defined as $P_{output} =[p_{d}, p_{n}]$, action control receives these two distributions and decides the final output action $a_t$}
	\label{DITA_arichitecture}
    \vspace{-10pt}
\end{figure}

\subsection{Method}

\textbf{Deep Reinforcement Learning Branch.}  Given the impressive capabilities of enriched environment exploration ability of MJOLNIR-o \citep{pal2021learning}, we use it as our backbone reinforcement learning model. Upon receiving the observation, the model builds a 2D array in shape $(N_C,N_C + 300)$ called Node Feature Matrix by processing the result from a ground-truth object detector, where $N_C = |C|$ is the number of object types across all rooms. Each row of the Node Feature Matrix would be passed as an individual input node feature pass to the corresponding GCN node, with its first $N_C$ columns standing for a binary vector indicating the object detection result for all object types $C$, and the last $300$ elements is a GloVe word embedding \citep{pennington-etal-2014-glove} vector of the current object. Node embedding is learned through a graph neural network that was made by object relation labels provided by Visual Genome (VG) dataset \citep{krishna2017visual} and pruned some relations off for AI2-THOR objects. On the other hand, the model also constructs Context Matrix from object detection, with each row representing a vector containing the object detection state of an object type $c \in C$ with $row_c = \{b,x_c,y_c,Bbx,CS\}$, b is a binary indicator represents whether an object with type $c$ is visible in the current frame, $x_c$ and $y_c$ is the coordinates of object detection bounding box center, $Bbx$ is the bounding box area, and the $CS$ is the cosine similarity of word embedding vectors between object type $c$ and the target object type, defined as:  
$
CS(G_{c},G_{target}) = \frac{G_{c} \cdot G_{target}}{||G_{c}|| \cdot ||G_{target}||}
$
. $G_{c}$ and $G_{target}$ are GloVe vectors for the current object and target object respectively.

Our evaluation of the environment points out that occasionally more than one instance of object type $c$ could be visible, \cite{pal2021learning} deals with this by averaging their bounding box center and area by default, but if two instances with the identical object type of large size show in one frame, the averaged bounding box might cover a lot of irrelevant smaller objects with other types. Moreover, since our judge model will receive information from the context matrix as input, such an approach leads to the problem of providing dirty data. In contrast, when multiple instances of type $c$ occur in the same frame, we take the one with the largest $Bbx$ to represent the class. The learned node embedding and the flattened context matrix are concatenated as joint embedding, passed to an LSTM cell, and sent to the A3C model to learn the control action distribution $P_{con}$.




\textbf{Judge Model Branch.} At each time step $t$, if $Done$ is sampled by the DRL branch, the judge model branch processes the flattened image feature $ImgEmb_t$ of the observation, extracted via a pre-trained ResNet-18 \citep{7780459} encoder. This encoder is pre-trained on ImageNet \citep{5206848}, encompassing 1000 object classes. By evaluating the context matrix obtained from the reinforcement learning branch, the judge model branch selects the target row with $CS = 1.0$ as the target state vector. The image features $ImgEmb_t$, target state vector $TagVec_t$ from the context matrix, and glove word embedding of the target $GloveEmb_t$ are concatenated to form a state embedding $StateEmb_t$. The judge model is trained only on \textit{Effective States} — states where the target is visible in the observation. If the target is not visible in the current frame (as indicated by $b=0$ in $StateEmb_t$), the current time step is ignored by the judge model, yielding no output. If the target is visible, $StateEmb_t$ is passed to the judge model. The output is then forwarded to the action control module. The agent acts on the final output action decided by the action control model and receives the reward signal. Analysis of the reward range reveals that successful episodes yield rewards in the range $R_t \in [4.05,4.90]$. If $R_t >= 4.0$, the ground truth for time step $t$ is set as positive; otherwise, it's set as negative. The ground truth of time step $t$ and the $StateEmb$ are stored as learning data in a "Batch Buffer" with a capacity of 64 samples. Upon reaching the maximum batch size, these samples serve as a training batch for the judge model to update the weights. This progress is illustrated in Figure \ref{DITA_arichitecture}.

\begin{figure}
	\centering
	\includegraphics[scale=0.35]{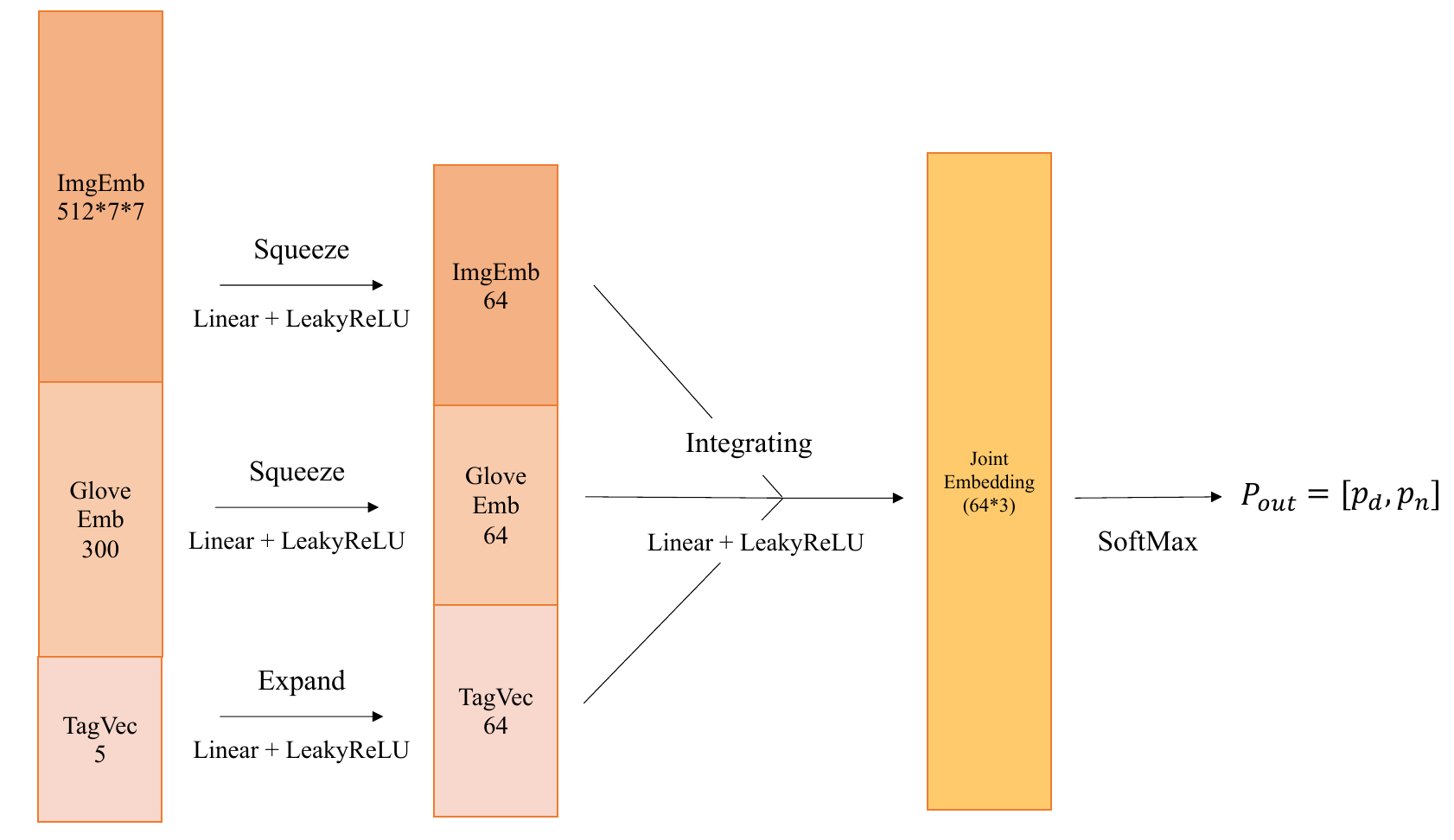}
	\caption{Judge Model. Adapt each component within \textit{StateEmb} to the same dimension, then fuse them as a joint embedding to learn termination classification.}
    
	\label{Judge Model}
    \vspace{-5pt}
\end{figure}

Judge model is a supervised binary classification neural network with expanding and squeezing layers, as shown in Figure \ref{Judge Model}, these layers map the input $StateEmb$ into the same dimension by several stacked linear layers, since $GloveEmb$ might contain negative floating numbers, we observe that applying ReLU activation after linear layers causes the gradient of large partition of neurons to be zero, therefore we use Leaky ReLU \citep{xu2015empirical} activation following the linear layers to prevent dead ReLU problem \citep{lu2019dying}. Eventually, concatenate $ImgEmb$, $GloveEmb$, and $TecVec$ together to form a joint embedding, and output the classification result with probabilities for whether to sample termination. In addition, because our data is collected online by reinforcement learning, during an episode, as mentioned in section \ref{sec:task}, since the success condition requires the agent to terminate within a certain range of the target, most of the \textit{Effective States} comes with ground truth of the negative class, where the agent should not terminate, this imbalance of training data causes long tail problem \citep{zhang2023deep}. In our method, we use Focal Loss proposed by \cite{lin2017focal} as our loss function as an alternative to Cross Entropy Loss:

\begin{equation}
    FL(p_t) = -(1-p_t)^\gamma log(p_t)
\end{equation}

Focal loss dynamically adjusts the weight of each instance in the loss function, focusing more on hard-to-classify instances and less on easy ones. We set $\gamma = 0.7$ in our experiments.

\begin{equation}
\label{action}
  a_t = \left\{
    \begin{array}{ll}
      Done, & \mbox{if $p_{d} + p_{\lambda} >= 1.5$}\\
      P_{con}, & \mbox{if $p_{d}$ is sampled} \\
      P_{sub}, & \mbox{if $p_{n}$ is sampled} \\
    \end{array}
  \right.
\end{equation}

\textbf{Action Control.} Action control directly samples the action from the output of reinforcement learning branch $P_{con}$ in training. However, in the testing phase, the action model relies on probability distributions generated by two models $P_{con}$ and $P_{out}$, and the action model outputs the final action $a_t = Done$ if both models express sufficient confidence in terminating the episode. 

Specifically, note the probability output of action \textit{Done} from $P_{con}$ as $p_{\lambda}$, and the probability of sample termination action in $P_{out}$ as $p_d$, output $a_t = Done$ when the sum of the confidence for termination action in two distributions satisfies $p_{d} + p_{\lambda} >= 1.5$. Otherwise, according to the output of the judge model, while $p_{d}$ is sampled from the output of the judge model, indicating that termination is advisable at the current time step, action control outputs final action $a_t \in P_{con}$. On the other hand, if $p_n$ is sampled by the judge model, suggesting that termination should be delayed, action control outputs final action $a_t \in P_{sub}$ with $P_{sub}$ being a subset of $P_{con}$ without \textit{Done} action. This decision process is formally represented in Equation \ref{action}.


\section{Experiment Results}
\label{sec:result}

\textbf{Environment \& Dataset.} We use AI2-THOR \citep{kolve2017ai2} as our environment simulator to evaluate our method for object navigation. AI2-THOR contains 120 different rooms with  30 rooms per room type Kitchen, Bedroom, Living room, and Bathroom. The rooms were split as training data and testing data, in our experiments, we use 80 rooms as training data, with 20 rooms from each room type. The remaining 40 rooms were used for testing. Amount all object categories in AI2-THOR environment $|C_{total}|=101$. 

\textbf{Evaluation Metrics.} The comparison of models was conducted using two metrics, in line with previous research~\citep{zhu2017target,wortsman2019learning}. \textit{Success Rate (SR)} measures the probability of agent success in the environment, computed by $SR=\frac{1}{N} \sum_{n=0}^N S_n$, $N$ is the number of total episodes in evaluation, and $S_n$ is a binary indicator 
 with $S_n=1$ represents agent succeed in episode $n$. In addition, we use \textit{Success Weighted
by Path Length (SPL)}, which measures the navigation efficiency of the agent, defined as $SPL=\frac{1}{N} \sum_{n=0}^N S_n  \frac{O_n}{max(L_n,O_n)}$ Where $O_n$ is the length of the optimal path to the target that agent could take in episode $n$, $L_n$ is the actual path length agent has taken.

\subsection{Compared Methods} 

We compare our method with other end-to-end reinforcement learning-based methods: \textbf{- Random} In a random model, the agent navigates in the environment by randomly sampled action. \textbf{- Target-driven VN} \citep{zhu2017target} Only fusions the observation of agent and the target embedding as input states to the model. \textbf{- Scene Prior} \citep{yang2018visual} This model incorporates semantic object relations as knowledge graph to the agent, learning from a joint embedding consisting of knowledge graph node embedding, image-wise observation features from pre-trained ResNet-18 and target word embedding. \textbf{- SAVN} \citep{wortsman2019learning} This model leverages meta-learning for the agent to learn the environment in both training and inferring.  \textbf{- MJOLNIR-o} \citep{pal2021learning} This model integrates hierarchical object relationships to the agent by reward shaping, and learning object-wise observation features by constructing a context matrix from an object detector. \textbf{- MJOLNIR-r} \citep{pal2021learning} MJOLNIR-r is an alternative version of the MJOLNIR-o model, which passes image-wise observation features to the agent rather than object-wise observation.



\begin{table}[]
\centering
\begin{tabular}{lccccc}

\hline & \\[-1.5ex]
            & \multicolumn{2}{c}{All}                                  & \multicolumn{1}{l}{} & \multicolumn{2}{c}{L \textgreater{}= 5}                  \\
            & \multicolumn{1}{l}{SR(\%)} & \multicolumn{1}{l}{SPL(\%)} & \multicolumn{1}{l}{} & \multicolumn{1}{l}{SR(\%)} & \multicolumn{1}{l}{SPL(\%)} \\ \cline{2-3} \cline{5-6} 
            \\[-1.5ex]
Random      & 10.4                       & 3.2                         &                      & 0.6                        & 0.4                         \\

Target-driven VN \citep{zhu2017target} & 35.0                       & 10.3                        &                      & 25.0                    & 10.5                     \\

Scene Prior \citep{yang2018visual} & 35.4                       & 10.9                        &                      & 23.8                       & 10.7                        \\
SAVN    \citep{wortsman2019learning}    & 35.7                       & 9.3                         &                      & 23.9                       & 9.4                         \\

MJOLNIR-r \citep{pal2021learning}  & 54.8                       & 19.2                      &                      & 41.7                      & 18.9                       \\

MJOLNIR-o \citep{pal2021learning}  & 65.3                       & 21.1                        &                      & 50.0                       & 20.9                        \\


\hline & \\[-1.5ex]

\textbf{DITA (Ours)} & \textbf{71.4}             & \textbf{21.6}                  &                      & \textbf{57.9}              & \textbf{22.2}     \\
    
\hline & \\[-1.5ex]

\end{tabular}
\caption{Experiment results with comparisons to other methods in AI2-THOR.}
\label{overallresult}
\vspace{-14pt}
\end{table}

\subsection{Results}

\begin{figure}[htbp]
 \begin{minipage}{0.45\textwidth}
  \centering
  \includegraphics[width=\textwidth]{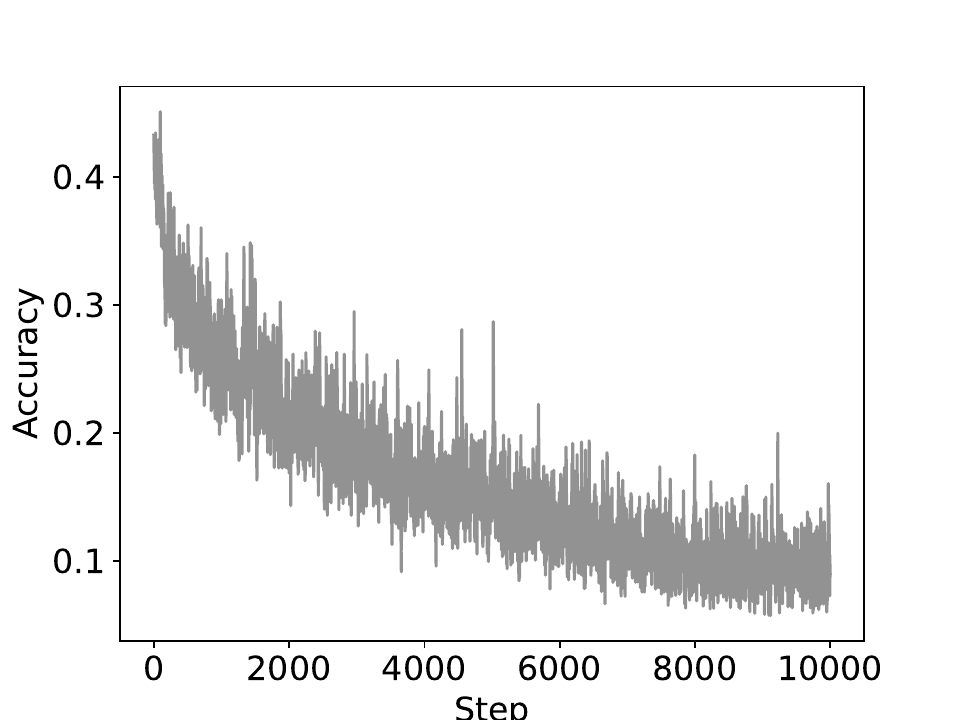}
  \caption{Training Loss of Judge Model with Smooth Factor $\beta = 0.8$.}
  \label{fig:lossfig}
 \end{minipage}
 \hfill
 \begin{minipage}{0.5\textwidth}
  \centering
  \includegraphics[width=\textwidth]{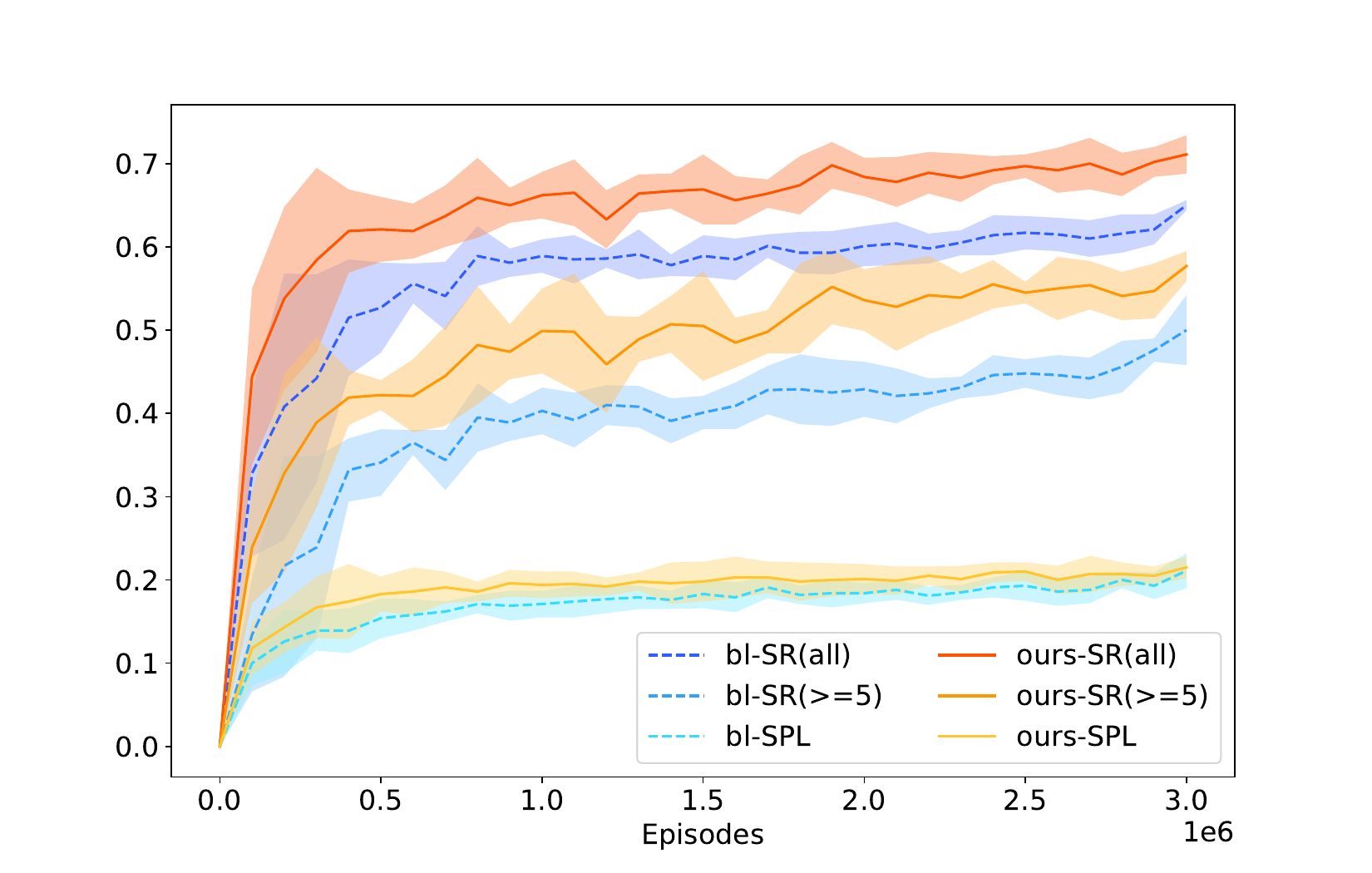}
  \caption{Test Accuracy of DITA and Baseline Model over 3 random seeds.}
  \label{fig:testaccu}
 \end{minipage}
\end{figure}

By constructing $StateEmb$ and passing it as input data, together with the reward signal, we have successfully trained a model to effectively recognize termination and handle termination action jointly with reinforcement learning, Figure \ref{fig:lossfig} illustrates the convergence of training loss of the judge model.

Compared with other map-less end-to-end models in Table \ref{overallresult} and Figure \ref{fig:testaccu}, it is evident that DITA demonstrates superior performance, particularly when compared to MJOLNIR-o. DITA exhibits a remarkable improvement of 9.3\% across all room types. Furthermore, DITA showcases a remarkable 15.8\% increase in episodes with optimal lengths ($L >= 5$). The significant improvements can be attributed to the novel architecture of DITA and its ability to implicitly infer object-wise depth information. This critical component helps to solve the problem of termination recognition in long episodes which other models struggle to handle effectively. By inferring depth information, DITA can better understand when the agent is close enough to the target object to end the episode successfully. This mechanism improves the success rate, especially in environments with long optimal episode lengths, Table \ref{perroomresult} offers a detailed performance breakdown of DITA and other models by room types. In environments with large size (hence larger episode length) and complex layouts like living rooms, which make navigation more challenging our DITA significantly surpasses all the other models in both metrics, achieving a success rate of 75.6\% and an SRL of 22.2\%. Our results demonstrate the effectiveness of considering episode termination separately for the deep reinforcement learning model. It is noticeable that in the case of the Bathroom environment of our result, MJOLNIR-o achieves the highest success rate (SR). Indicates that DITA is capable of effectively navigating more confined and object-dense environments. We observed modest improvements in  Success Weighted by Path Length (SRL). These results underscore the challenges involved in path planning and termination recognition in such scenarios.

\begin{table}[t]
\resizebox{\textwidth}{!}{%
\renewcommand{\arraystretch}{1.3}
\begin{tabular}{l cc cc cc cc cc}
\toprule
\multicolumn{1}{c}{\multirow{2}{*}{\textbf{Model}}} & \multicolumn{2}{c}{\textbf{Bath Room}} & \multicolumn{2}{c}{\textbf{Bedroom}} & \multicolumn{2}{c}{\textbf{Kitchen}} & \multicolumn{2}{c}{\textbf{Living Room}} & \multicolumn{2}{c}{\textbf{Avg.}} \\
\multicolumn{1}{c}{}                                & SR(\%)             & SRL(\%)            & SR(\%)            & SRL(\%)           & SR(\%)            & SRL(\%)           & SR(\%)              & SRL(\%)             & SR(\%)           & SRL(\%)         \\ \hline 
Target-driven VN \citep{zhu2017target}                                     & 53.2               & 13.4               & 28.8              & 9.0               & 32.4              & 10.9              & 35.2                & 10.0                & 37.4             & 10.8            \\
Scene Prior \citep{yang2018visual}                                           & 41.6               & 13.3               & 33.6              & 10.4              & 26.4              & 9.1               & 36.0                & 9.9                 & 34.4             & 10.7            \\
SAVN \citep{wortsman2019learning}                                                  & 47.6               & 14.6               & 21.6              & 6.7               & 34.8              & 8.3               & 40.0                & 9.0                 & 36.9             & 9.7             \\
MJOLNIR-r  \citep{pal2021learning}                                           & 72.8               & 24.3               & 41.2              & 16.9              & 56.4              & 21.2              & 50.8                & 15.9                & 55.3             & 19.6            \\
MJOLNIR-o  \citep{pal2021learning}                                           & \textbf{82.4}      & \textbf{25.1}      & 43.2              & 14.4              & \textbf{74.8}     & 22.9     & 50.0                & 17.9                & 62.6             & 20.1            \\  \cmidrule{1-11}
\textbf{DITA (Ours)}                                  & 63.2               & 20.1               & \textbf{61.5}     & \textbf{18.6}     & 73.0              & \textbf{ 23.0 }             & \textbf{75.6}       & \textbf{22.2}       & \textbf{68.3}    & \textbf{21.0}   \\ \hline

\end{tabular}
}
\caption{Experiment results by room types.}
\label{perroomresult}
\end{table}

\section{Discussions}
\label{sec:conclusion}

\subsection{Limitations and Future Work}

We have conducted failure cases analysis for DITA model, mainly the agent fails in the following cases: (1) Target object needs a precise path to navigate to. For example, a pillow of a double bed in a narrow room, where an agent needs to navigate precisely to the front corner of the bed, indicates the agent might still need an explicit planning component. (2) We observe that the training time required for training DITA is dominating all other tested models, involving some efficient methods to the architecture might even boost the performance.

\subsection{Conclusion}

This paper presents the Depth-Inference Termination Agent (DITA), a novel approach designed to tackle the challenge of object navigation in autonomous navigation systems. Focusing specifically on the issues of target approach and episode termination in environments with lengthy optimal episode length, our approach has shown promising results in overcoming limitations faced by conventional Deep Reinforcement Learning (DRL) methods. Our experimental results, conducted within the AI2-THOR framework, clearly illustrate the superior performance of DITA.  Our experiment results also highlight opportunities for further enhancements, possibly through the refinement of depth estimation, exploration strategies, or incorporation of additional environmental cues.


\bibliography{acml23}

\begin{thebibliography}{51}
\providecommand{\natexlab}[1]{#1}
\providecommand{\url}[1]{\texttt{#1}}
\expandafter\ifx\csname urlstyle\endcsname\relax
  \providecommand{\doi}[1]{doi: #1}\else
  \providecommand{\doi}{doi: \begingroup \urlstyle{rm}\Url}\fi

\bibitem[Andrychowicz et~al.(2017)Andrychowicz, Wolski, Ray, Schneider, Fong,
  Welinder, McGrew, Tobin, Pieter~Abbeel, and
  Zaremba]{andrychowicz2017hindsight}
Marcin Andrychowicz, Filip Wolski, Alex Ray, Jonas Schneider, Rachel Fong,
  Peter Welinder, Bob McGrew, Josh Tobin, OpenAI Pieter~Abbeel, and Wojciech
  Zaremba.
\newblock Hindsight experience replay.
\newblock \emph{Advances in neural information processing systems}, 30, 2017.

\bibitem[Bagnell et~al.(2010)Bagnell, Bradley, Silver, Sofman, and
  Stentz]{bagnell2010learning}
James~Andrew Bagnell, David Bradley, David Silver, Boris Sofman, and Anthony
  Stentz.
\newblock Learning for autonomous navigation.
\newblock \emph{IEEE Robotics \& Automation Magazine}, 17\penalty0
  (2):\penalty0 74--84, 2010.

\bibitem[Chaplot et~al.(2020{\natexlab{a}})Chaplot, Gandhi, Gupta, Gupta, and
  Salakhutdinov]{chaplot2020learning}
Devendra~Singh Chaplot, Dhiraj Gandhi, Saurabh Gupta, Abhinav Gupta, and Ruslan
  Salakhutdinov.
\newblock Learning to explore using active neural slam.
\newblock \emph{arXiv preprint arXiv:2004.05155}, 2020{\natexlab{a}}.

\bibitem[Chaplot et~al.(2020{\natexlab{b}})Chaplot, Gandhi, Gupta, and
  Salakhutdinov]{chaplot2020object}
Devendra~Singh Chaplot, Dhiraj~Prakashchand Gandhi, Abhinav Gupta, and Russ~R
  Salakhutdinov.
\newblock Object goal navigation using goal-oriented semantic exploration.
\newblock \emph{Advances in Neural Information Processing Systems},
  33:\penalty0 4247--4258, 2020{\natexlab{b}}.

\bibitem[Deng et~al.(2009)Deng, Dong, Socher, Li, Li, and Fei-Fei]{5206848}
Jia Deng, Wei Dong, Richard Socher, Li-Jia Li, Kai Li, and Li~Fei-Fei.
\newblock Imagenet: A large-scale hierarchical image database.
\newblock In \emph{2009 IEEE Conference on Computer Vision and Pattern
  Recognition}, pages 248--255, 2009.
\newblock \doi{10.1109/CVPR.2009.5206848}.

\bibitem[Devlin and Kudenko(2012)]{devlin2012dynamic}
Sam~Michael Devlin and Daniel Kudenko.
\newblock Dynamic potential-based reward shaping.
\newblock In \emph{Proceedings of the 11th international conference on
  autonomous agents and multiagent systems}, pages 433--440. IFAAMAS, 2012.

\bibitem[Dinh~Vuong et~al.(2023)Dinh~Vuong, Tien~Nguyen, Nhat~VU, Huang,
  Nguyen, Thanh~Binh, Vo, and Nguyen]{dinh2023habicrowd}
An~Dinh~Vuong, Toan Tien~Nguyen, Minh Nhat~VU, Baoru Huang, Dzung Nguyen,
  Huynh~Thi Thanh~Binh, Thieu Vo, and Anh Nguyen.
\newblock Habicrowd: A high performance simulator for crowd-aware visual
  navigation.
\newblock \emph{arXiv e-prints}, pages arXiv--2306, 2023.

\bibitem[Druon et~al.(2020)Druon, Yoshiyasu, Kanezaki, and
  Watt]{druon2020visual}
Raphael Druon, Yusuke Yoshiyasu, Asako Kanezaki, and Alassane Watt.
\newblock Visual object search by learning spatial context.
\newblock \emph{IEEE Robotics and Automation Letters}, 5\penalty0 (2):\penalty0
  1279--1286, 2020.

\bibitem[Du et~al.(2020)Du, Yu, and Zheng]{du2020learning}
Heming Du, Xin Yu, and Liang Zheng.
\newblock Learning object relation graph and tentative policy for visual
  navigation.
\newblock In \emph{Computer Vision--ECCV 2020: 16th European Conference,
  Glasgow, UK, August 23--28, 2020, Proceedings, Part VII 16}, pages 19--34.
  Springer, 2020.

\bibitem[Du et~al.(2021)Du, Yu, and Zheng]{du2021vtnet}
Heming Du, Xin Yu, and Liang Zheng.
\newblock Vtnet: Visual transformer network for object goal navigation.
\newblock \emph{arXiv preprint arXiv:2105.09447}, 2021.

\bibitem[Fuentes-Pacheco et~al.(2015)Fuentes-Pacheco, Ruiz-Ascencio, and
  Rend{\'o}n-Mancha]{fuentes2015visual}
Jorge Fuentes-Pacheco, Jos{\'e} Ruiz-Ascencio, and Juan~Manuel
  Rend{\'o}n-Mancha.
\newblock Visual simultaneous localization and mapping: a survey.
\newblock \emph{Artificial intelligence review}, 43:\penalty0 55--81, 2015.

\bibitem[Fukushima et~al.(2022)Fukushima, Ota, Kanezaki, Sasaki, and
  Yoshiyasu]{9812027}
Rui Fukushima, Kei Ota, Asako Kanezaki, Yoko Sasaki, and Yusuke Yoshiyasu.
\newblock Object memory transformer for object goal navigation.
\newblock In \emph{2022 International Conference on Robotics and Automation
  (ICRA)}, pages 11288--11294, 2022.
\newblock \doi{10.1109/ICRA46639.2022.9812027}.

\bibitem[Haarnoja et~al.(2018)Haarnoja, Zhou, Abbeel, and
  Levine]{haarnoja2018soft}
Tuomas Haarnoja, Aurick Zhou, Pieter Abbeel, and Sergey Levine.
\newblock Soft actor-critic: Off-policy maximum entropy deep reinforcement
  learning with a stochastic actor.
\newblock In \emph{International conference on machine learning}, pages
  1861--1870. PMLR, 2018.

\bibitem[He et~al.(2016)He, Zhang, Ren, and Sun]{7780459}
Kaiming He, Xiangyu Zhang, Shaoqing Ren, and Jian Sun.
\newblock Deep residual learning for image recognition.
\newblock In \emph{2016 IEEE Conference on Computer Vision and Pattern
  Recognition (CVPR)}, pages 770--778, 2016.
\newblock \doi{10.1109/CVPR.2016.90}.

\bibitem[Ho and Ermon(2016)]{ho2016generative}
Jonathan Ho and Stefano Ermon.
\newblock Generative adversarial imitation learning.
\newblock \emph{Advances in neural information processing systems}, 29, 2016.

\bibitem[Hu et~al.(2020)Hu, Wang, Jia, Wang, Chen, Hao, Wu, and
  Fan]{hu2020learning}
Yujing Hu, Weixun Wang, Hangtian Jia, Yixiang Wang, Yingfeng Chen, Jianye Hao,
  Feng Wu, and Changjie Fan.
\newblock Learning to utilize shaping rewards: A new approach of reward
  shaping.
\newblock \emph{Advances in Neural Information Processing Systems},
  33:\penalty0 15931--15941, 2020.

\bibitem[Jaakkola et~al.(1994)Jaakkola, Singh, and
  Jordan]{jaakkola1994reinforcement}
Tommi Jaakkola, Satinder Singh, and Michael Jordan.
\newblock Reinforcement learning algorithm for partially observable markov
  decision problems.
\newblock \emph{Advances in neural information processing systems}, 7, 1994.

\bibitem[Kartal et~al.(2019)Kartal, Hernandez-Leal, and
  Taylor]{kartal2019terminal}
Bilal Kartal, Pablo Hernandez-Leal, and Matthew~E Taylor.
\newblock Terminal prediction as an auxiliary task for deep reinforcement
  learning.
\newblock In \emph{Proceedings of the AAAI Conference on Artificial
  Intelligence and Interactive Digital Entertainment}, volume~15, pages 38--44,
  2019.

\bibitem[Khandelwal et~al.(2022)Khandelwal, Weihs, Mottaghi, and
  Kembhavi]{khandelwal2022simple}
Apoorv Khandelwal, Luca Weihs, Roozbeh Mottaghi, and Aniruddha Kembhavi.
\newblock Simple but effective: Clip embeddings for embodied ai.
\newblock In \emph{Proceedings of the IEEE/CVF Conference on Computer Vision
  and Pattern Recognition}, pages 14829--14838, 2022.

\bibitem[Kolve et~al.(2017)Kolve, Mottaghi, Han, VanderBilt, Weihs, Herrasti,
  Deitke, Ehsani, Gordon, Zhu, et~al.]{kolve2017ai2}
Eric Kolve, Roozbeh Mottaghi, Winson Han, Eli VanderBilt, Luca Weihs, Alvaro
  Herrasti, Matt Deitke, Kiana Ehsani, Daniel Gordon, Yuke Zhu, et~al.
\newblock Ai2-thor: An interactive 3d environment for visual ai.
\newblock \emph{arXiv preprint arXiv:1712.05474}, 2017.

\bibitem[Krishna et~al.(2017)Krishna, Zhu, Groth, Johnson, Hata, Kravitz, Chen,
  Kalantidis, Li, Shamma, et~al.]{krishna2017visual}
Ranjay Krishna, Yuke Zhu, Oliver Groth, Justin Johnson, Kenji Hata, Joshua
  Kravitz, Stephanie Chen, Yannis Kalantidis, Li-Jia Li, David~A Shamma, et~al.
\newblock Visual genome: Connecting language and vision using crowdsourced
  dense image annotations.
\newblock \emph{International journal of computer vision}, 123:\penalty0
  32--73, 2017.

\bibitem[Laina et~al.(2016)Laina, Rupprecht, Belagiannis, Tombari, and
  Navab]{laina2016deeper}
Iro Laina, Christian Rupprecht, Vasileios Belagiannis, Federico Tombari, and
  Nassir Navab.
\newblock Deeper depth prediction with fully convolutional residual networks.
\newblock In \emph{2016 Fourth international conference on 3D vision (3DV)},
  pages 239--248. IEEE, 2016.

\bibitem[Lin et~al.(2017)Lin, Goyal, Girshick, He, and
  Doll{\'a}r]{lin2017focal}
Tsung-Yi Lin, Priya Goyal, Ross Girshick, Kaiming He, and Piotr Doll{\'a}r.
\newblock Focal loss for dense object detection.
\newblock In \emph{Proceedings of the IEEE international conference on computer
  vision}, pages 2980--2988, 2017.

\bibitem[Lu et~al.(2019)Lu, Shin, Su, and Karniadakis]{lu2019dying}
Lu~Lu, Yeonjong Shin, Yanhui Su, and George~Em Karniadakis.
\newblock Dying relu and initialization: Theory and numerical examples.
\newblock \emph{arXiv preprint arXiv:1903.06733}, 2019.

\bibitem[Milani et~al.(2023)Milani, Juliani, Momennejad, Georgescu, Rzepecki,
  Shaw, Costello, Fang, Devlin, and Hofmann]{milani2023navigates}
Stephanie Milani, Arthur Juliani, Ida Momennejad, Raluca Georgescu, Jaroslaw
  Rzepecki, Alison Shaw, Gavin Costello, Fei Fang, Sam Devlin, and Katja
  Hofmann.
\newblock Navigates like me: Understanding how people evaluate human-like ai in
  video games.
\newblock In \emph{Proceedings of the 2023 CHI Conference on Human Factors in
  Computing Systems}, pages 1--18, 2023.

\bibitem[Mirowski et~al.(2016)Mirowski, Pascanu, Viola, Soyer, Ballard, Banino,
  Denil, Goroshin, Sifre, Kavukcuoglu, et~al.]{mirowski2016learning}
Piotr Mirowski, Razvan Pascanu, Fabio Viola, Hubert Soyer, Andrew~J Ballard,
  Andrea Banino, Misha Denil, Ross Goroshin, Laurent Sifre, Koray Kavukcuoglu,
  et~al.
\newblock Learning to navigate in complex environments.
\newblock \emph{arXiv preprint arXiv:1611.03673}, 2016.

\bibitem[Mnih et~al.(2016)Mnih, Badia, Mirza, Graves, Lillicrap, Harley,
  Silver, and Kavukcuoglu]{mnih2016asynchronous}
Volodymyr Mnih, Adria~Puigdomenech Badia, Mehdi Mirza, Alex Graves, Timothy
  Lillicrap, Tim Harley, David Silver, and Koray Kavukcuoglu.
\newblock Asynchronous methods for deep reinforcement learning.
\newblock In \emph{International conference on machine learning}, pages
  1928--1937. PMLR, 2016.

\bibitem[Oriolo et~al.(1995)Oriolo, Vendittelli, and Ulivi]{oriolo1995line}
Giuseppe Oriolo, Marilena Vendittelli, and Giovanni Ulivi.
\newblock On-line map building and navigation for autonomous mobile robots.
\newblock In \emph{Proceedings of 1995 IEEE international conference on
  robotics and automation}, volume~3, pages 2900--2906. IEEE, 1995.

\bibitem[Ostrovski et~al.(2017)Ostrovski, Bellemare, Oord, and
  Munos]{ostrovski2017count}
Georg Ostrovski, Marc~G Bellemare, A{\"a}ron Oord, and R{\'e}mi Munos.
\newblock Count-based exploration with neural density models.
\newblock In \emph{International conference on machine learning}, pages
  2721--2730. PMLR, 2017.

\bibitem[Pal et~al.(2021)Pal, Qiu, and Christensen]{pal2021learning}
Anwesan Pal, Yiding Qiu, and Henrik Christensen.
\newblock Learning hierarchical relationships for object-goal navigation.
\newblock In \emph{Conference on Robot Learning}, pages 517--528. PMLR, 2021.

\bibitem[Pathak et~al.(2017)Pathak, Agrawal, Efros, and
  Darrell]{pathak2017curiosity}
Deepak Pathak, Pulkit Agrawal, Alexei~A Efros, and Trevor Darrell.
\newblock Curiosity-driven exploration by self-supervised prediction.
\newblock In \emph{International conference on machine learning}, pages
  2778--2787. PMLR, 2017.

\bibitem[Pennington et~al.(2014)Pennington, Socher, and
  Manning]{pennington-etal-2014-glove}
Jeffrey Pennington, Richard Socher, and Christopher Manning.
\newblock {G}lo{V}e: Global vectors for word representation.
\newblock In \emph{Proceedings of the 2014 Conference on Empirical Methods in
  Natural Language Processing ({EMNLP})}, pages 1532--1543, Doha, Qatar,
  October 2014. Association for Computational Linguistics.
\newblock \doi{10.3115/v1/D14-1162}.
\newblock URL \url{https://aclanthology.org/D14-1162}.

\bibitem[Ramakrishnan et~al.(2022)Ramakrishnan, Chaplot, Al-Halah, Malik, and
  Grauman]{ramakrishnan2022poni}
Santhosh~Kumar Ramakrishnan, Devendra~Singh Chaplot, Ziad Al-Halah, Jitendra
  Malik, and Kristen Grauman.
\newblock Poni: Potential functions for objectgoal navigation with
  interaction-free learning.
\newblock In \emph{Proceedings of the IEEE/CVF Conference on Computer Vision
  and Pattern Recognition}, pages 18890--18900, 2022.

\bibitem[Ramrakhya et~al.(2022)Ramrakhya, Undersander, Batra, and
  Das]{ramrakhya2022habitat}
Ram Ramrakhya, Eric Undersander, Dhruv Batra, and Abhishek Das.
\newblock Habitat-web: Learning embodied object-search strategies from human
  demonstrations at scale.
\newblock In \emph{Proceedings of the IEEE/CVF Conference on Computer Vision
  and Pattern Recognition}, pages 5173--5183, 2022.

\bibitem[Ranjan et~al.(2019)Ranjan, Jampani, Balles, Kim, Sun, Wulff, and
  Black]{ranjan2019competitive}
Anurag Ranjan, Varun Jampani, Lukas Balles, Kihwan Kim, Deqing Sun, Jonas
  Wulff, and Michael~J Black.
\newblock Competitive collaboration: Joint unsupervised learning of depth,
  camera motion, optical flow and motion segmentation.
\newblock In \emph{Proceedings of the IEEE/CVF conference on computer vision
  and pattern recognition}, pages 12240--12249, 2019.

\bibitem[Ren et~al.(2022)Ren, Huang, and Huang]{electronics11213628}
Jing Ren, Xishi Huang, and Raymond~N. Huang.
\newblock Efficient deep reinforcement learning for optimal path planning.
\newblock \emph{Electronics}, 11\penalty0 (21), 2022.
\newblock ISSN 2079-9292.
\newblock \doi{10.3390/electronics11213628}.
\newblock URL \url{https://www.mdpi.com/2079-9292/11/21/3628}.

\bibitem[Savva et~al.(2017)Savva, Chang, Dosovitskiy, Funkhouser, and
  Koltun]{savva2017minos}
Manolis Savva, Angel~X Chang, Alexey Dosovitskiy, Thomas Funkhouser, and
  Vladlen Koltun.
\newblock Minos: Multimodal indoor simulator for navigation in complex
  environments.
\newblock \emph{arXiv preprint arXiv:1712.03931}, 2017.

\bibitem[Soviany et~al.(2022)Soviany, Ionescu, Rota, and
  Sebe]{soviany2022curriculum}
Petru Soviany, Radu~Tudor Ionescu, Paolo Rota, and Nicu Sebe.
\newblock Curriculum learning: A survey.
\newblock \emph{International Journal of Computer Vision}, 130\penalty0
  (6):\penalty0 1526--1565, 2022.

\bibitem[Stadie et~al.(2015)Stadie, Levine, and
  Abbeel]{stadie2015incentivizing}
Bradly~C Stadie, Sergey Levine, and Pieter Abbeel.
\newblock Incentivizing exploration in reinforcement learning with deep
  predictive models.
\newblock \emph{arXiv preprint arXiv:1507.00814}, 2015.

\bibitem[Wang et~al.(2022)Wang, Zhang, Zhang, Fang, Xia, Liu, and
  Dong]{s22062387}
Fan Wang, Chaofan Zhang, Wen Zhang, Cuiyun Fang, Yingwei Xia, Yong Liu, and Hao
  Dong.
\newblock Object-based reliable visual navigation for mobile robot.
\newblock \emph{Sensors}, 22\penalty0 (6), 2022.
\newblock ISSN 1424-8220.
\newblock \doi{10.3390/s22062387}.
\newblock URL \url{https://www.mdpi.com/1424-8220/22/6/2387}.

\bibitem[Wijmans et~al.(2023)Wijmans, Savva, Essa, Lee, Morcos, and
  Batra]{wijmans2023emergence}
Erik Wijmans, Manolis Savva, Irfan Essa, Stefan Lee, Ari~S Morcos, and Dhruv
  Batra.
\newblock Emergence of maps in the memories of blind navigation agents.
\newblock \emph{arXiv preprint arXiv:2301.13261}, 2023.

\bibitem[Wortsman et~al.(2019)Wortsman, Ehsani, Rastegari, Farhadi, and
  Mottaghi]{wortsman2019learning}
Mitchell Wortsman, Kiana Ehsani, Mohammad Rastegari, Ali Farhadi, and Roozbeh
  Mottaghi.
\newblock Learning to learn how to learn: Self-adaptive visual navigation using
  meta-learning.
\newblock In \emph{Proceedings of the IEEE/CVF conference on computer vision
  and pattern recognition}, pages 6750--6759, 2019.

\bibitem[Xu et~al.(2015)Xu, Wang, Chen, and Li]{xu2015empirical}
Bing Xu, Naiyan Wang, Tianqi Chen, and Mu~Li.
\newblock Empirical evaluation of rectified activations in convolutional
  network.
\newblock \emph{arXiv preprint arXiv:1505.00853}, 2015.

\bibitem[Yang et~al.(2018)Yang, Wang, Farhadi, Gupta, and
  Mottaghi]{yang2018visual}
Wei Yang, Xiaolong Wang, Ali Farhadi, Abhinav Gupta, and Roozbeh Mottaghi.
\newblock Visual semantic navigation using scene priors.
\newblock \emph{arXiv preprint arXiv:1810.06543}, 2018.

\bibitem[Ye et~al.(2021)Ye, Batra, Das, and Wijmans]{ye2021auxiliary}
Joel Ye, Dhruv Batra, Abhishek Das, and Erik Wijmans.
\newblock Auxiliary tasks and exploration enable objectnav.
\newblock \emph{arXiv preprint arXiv:2104.04112}, 2021.

\bibitem[Zhang et~al.(2021)Zhang, Song, Bai, Li, Chu, and
  Jiang]{zhang2021hierarchical}
Sixian Zhang, Xinhang Song, Yubing Bai, Weijie Li, Yakui Chu, and Shuqiang
  Jiang.
\newblock Hierarchical object-to-zone graph for object navigation.
\newblock In \emph{Proceedings of the IEEE/CVF international conference on
  computer vision}, pages 15130--15140, 2021.

\bibitem[Zhang et~al.(2023)Zhang, Kang, Hooi, Yan, and Feng]{zhang2023deep}
Yifan Zhang, Bingyi Kang, Bryan Hooi, Shuicheng Yan, and Jiashi Feng.
\newblock Deep long-tailed learning: A survey.
\newblock \emph{IEEE Transactions on Pattern Analysis and Machine
  Intelligence}, 2023.

\bibitem[Zheng et~al.(2018)Zheng, Cham, and Cai]{zheng2018t2net}
Chuanxia Zheng, Tat-Jen Cham, and Jianfei Cai.
\newblock T2net: Synthetic-to-realistic translation for solving single-image
  depth estimation tasks.
\newblock In \emph{Proceedings of the European conference on computer vision
  (ECCV)}, pages 767--783, 2018.

\bibitem[Zhou et~al.(2017)Zhou, Brown, Snavely, and Lowe]{zhou2017unsupervised}
Tinghui Zhou, Matthew Brown, Noah Snavely, and David~G Lowe.
\newblock Unsupervised learning of depth and ego-motion from video.
\newblock In \emph{Proceedings of the IEEE conference on computer vision and
  pattern recognition}, pages 1851--1858, 2017.

\bibitem[Zhu et~al.(2021)Zhu, Liang, Zhu, Yu, Chang, and Liang]{zhu2021soon}
Fengda Zhu, Xiwen Liang, Yi~Zhu, Qizhi Yu, Xiaojun Chang, and Xiaodan Liang.
\newblock Soon: Scenario oriented object navigation with graph-based
  exploration.
\newblock In \emph{Proceedings of the IEEE/CVF Conference on Computer Vision
  and Pattern Recognition}, pages 12689--12699, 2021.

\bibitem[Zhu et~al.(2017)Zhu, Mottaghi, Kolve, Lim, Gupta, Fei-Fei, and
  Farhadi]{zhu2017target}
Yuke Zhu, Roozbeh Mottaghi, Eric Kolve, Joseph~J Lim, Abhinav Gupta,
  Li~Fei-Fei, and Ali Farhadi.
\newblock Target-driven visual navigation in indoor scenes using deep
  reinforcement learning.
\newblock In \emph{2017 IEEE international conference on robotics and
  automation (ICRA)}, pages 3357--3364. IEEE, 2017.

\end{thebibliography}

\appendix

\section{Reward Supervised Parallel Training}
\label{apd::trainingpro}

\begin{figure}[H]
	\centering
	\includegraphics[scale=0.42]{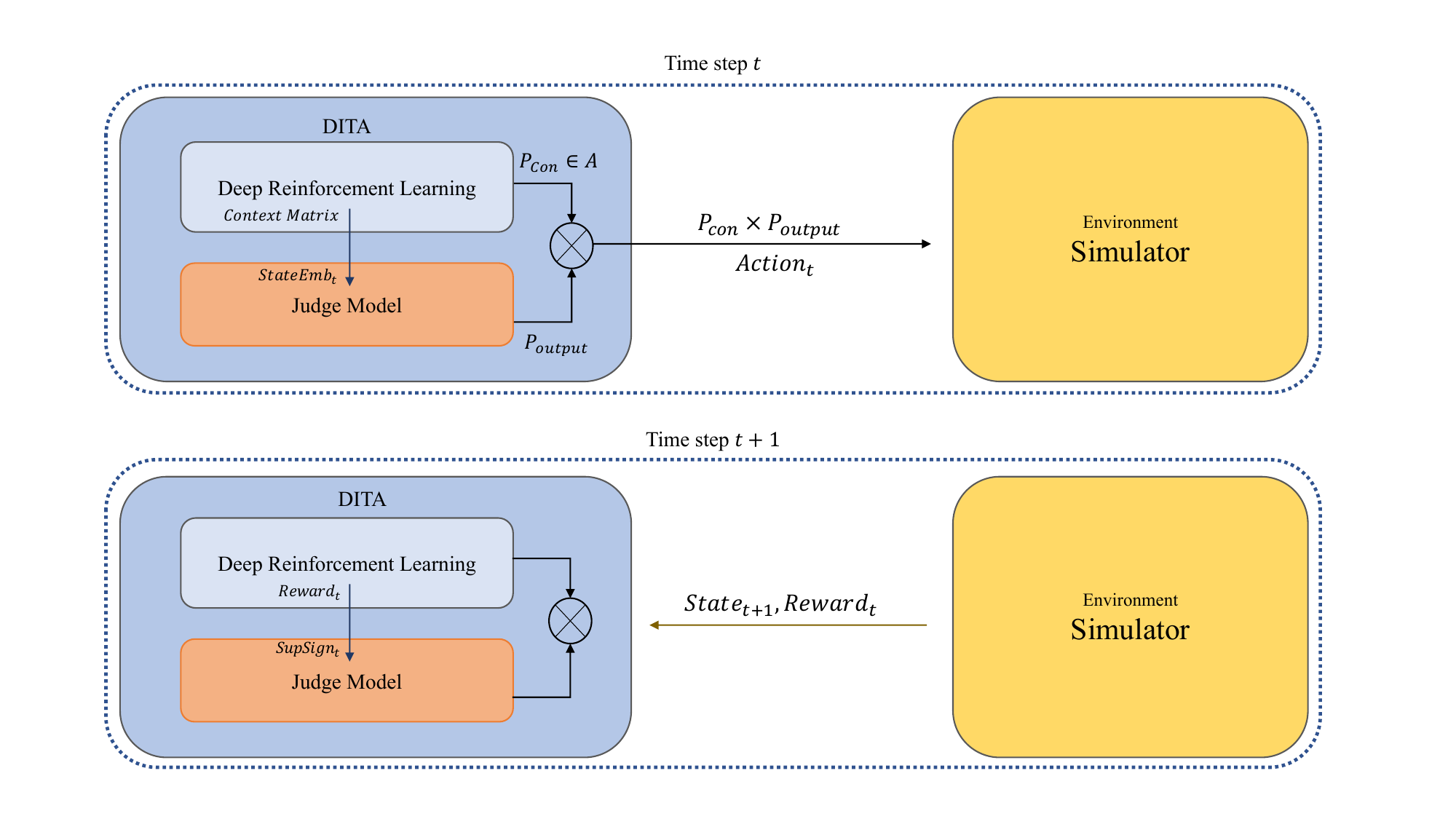}
	\caption{Reward Supervised Parallel Training.}
	\label{DITA_training}
\end{figure}

We train our reinforcement learning model and judge model in parallel, at time step $t$, the reinforcement learning model outputs control action distribution $P_{con}$, and the judge model outputs termination action distribution $P_{out}$ given $StateEmb_t$ from $Context \ Matrix$. Action control receives two outputs and decides the final action $a_t$, then in time step $t+1$, the reinforcement learning agent learns by the reward $Reward_{t}$ returned from the environment, whereas judge model transfers $Reward_{t}$ into ground truth supervision signal $SupSign_t$ and store it with $StateEmb_t$ as a sample data in \textit{Batch Buffer}, and updates itself once every $64$ sample were collected. Figure \ref{DITA_training} demonstrates this progress. 

\section{Target Object List}

\begin{table}[H]
\centering
\begin{tabular}{c  c}

\hline\\[-1.5ex]
\textbf{Room} & \textbf{Possible Target Object}                             \\ 

\\[-1.5ex]

Kitchen       & Toaster, Spatula, Bread, Mug, CoffeeMachine, Apple          \\ 

\\[-1.5ex]

Living room   & Painting, Laptop, Television, RemoteControl, Vase, ArmChair \\ 

\\[-1.5ex]

Bedroom       & Blinds, DeskLamp, Pillow, AlarmClock, CD                    \\ 

\\[-1.5ex]

Bathroom      & Mirror, ToiletPaper, SoapBar, Towel, SprayBottle            \\ 

\\[-1.5ex]\hline
\end{tabular}

\caption{List of Target Objects}
\end{table}

\section{Implementation Details}

We concurrently trained our judge model branch and the reinforcement learning branch with initially 1.6M episodes until we empirically observed that the judge model's accuracy had saturated, we then froze the judge model branch and continued to train the reinforcement learning branch with in total of 3.0M episodes for all models. Our training/testing division is consistent with \cite{pal2021learning,wortsman2019learning}. Models were trained on offline data collected from AI2-THOR v1.0.1. The A3C algorithm used in models was trained on 8 workers. 

\end{document}